\documentclass[runningheads]{llncs}

\usepackage[T1]{fontenc}
\usepackage{amsmath,amssymb,amsfonts}
\usepackage{graphicx}
\usepackage{textcomp}
\usepackage{xcolor}
\usepackage{url}
\usepackage{tabularx}
\usepackage{booktabs}
\usepackage{makecell}
\usepackage{float}
\usepackage{multirow}
\usepackage{caption}
\usepackage{subcaption}
\usepackage{pifont}
\usepackage{placeins}
\usepackage{arydshln}
\usepackage{array}
\usepackage{bm}
\usepackage[hidelinks,breaklinks=true]{hyperref}
\usepackage{capt-of}
\usepackage[ruled,vlined,linesnumbered]{algorithm2e}
\SetKwComment{Comment}{$\triangleright$\ }{}

\graphicspath{{../reference_material/figures/}}

\definecolor{navy}{RGB}{0,0,128}
\definecolor{darkgreen}{rgb}{0,0.5,0}

\begin{document}

\title{BEACON: Belief-Enabled Adaptive CONtrol\\
for Imitation Learning under Uncertainty}

\titlerunning{BEACON: Belief-Enabled Adaptive CONtrol}

\author{Moonyoung Lee\orcidID{0000-0002-1057-4556} , Soumojit Bhattacharya\orcidID{0009-0008-0380-6022} , George Kantor\orcidID{0000-0001-7088-853} , Oliver Kroemer\orcidID{0000-0003-2007-3867} }
\authorrunning{M. Lee et al.}
\institute{Carnegie Mellon University, Robotics Institute, Pittsburgh PA 15213, USA
\email{\{moonyoul,soumojib, gkantor, okroemer\}@andrew.cmu.edu}
}
\maketitle

\begin{abstract}
Robot manipulation tasks often involve hidden state information that cannot be directly observed and must be inferred through sequential physical interactions.
In such partially observable settings, conditioning an imitation learning policy directly on the recent raw observation history leads to poor performance. This is due to state aliasing, wherein identical observations may arise from different hidden states, and the policy receives conflicting action labels for the same input.
To enable history-aware disambiguation capability, we propose conditioning a diffusion policy on a structured representation of the hidden states using Bayesian belief $b_t = P(\mathbf{s} \mid z_{1:t})$ that exposes both the current most likely state estimate and the remaining uncertainty.
This representation replaces raw history with a structured, compact input, enabling the policy to implicitly modulate between exploratory and exploitative behaviors based on belief uncertainty, without explicit mode switching or reward shaping.
We evaluate across two domains with qualitatively different belief representations: a continuous belief for cornstalk gripper alignment via tactile sensing, and a discrete categorical distribution for latched door opening.
In both domains, the belief-conditioned policy substantially outperforms the observation-only baseline and approaches privileged ground-truth performance, with ablations illustrating that the policy adapts its exploration behavior depending on the belief uncertainty at inference time.

\keywords{Interactive Perception  \and Bayesian Belief \and Diffusion policy }
\end{abstract}

\section{Introduction}~\label{sec:intro}

Many robot manipulation tasks involve hidden state information that must be inferred through physical interaction rather than direct observation~\cite{wu2025savor,allevato2020tunenet,xu2019densephysnet}.
In agricultural robotics, for example, autonomous sensor insertion into plant stems~\cite{lee2024towards} requires estimating the stalk pose, yet dense foliage occludes visual estimates, making tactile sensing a critical complementary sensor modality.
A single tactile contact observation only partially disambiguates the hidden state, so the robot must make a sequence of intentional contacts to progressively resolve uncertainty before committing to task completion by aligning the gripper with the estimated stalk.
The hidden state need not be a continuous variable such as object pose; it can also be a discrete property, such as whether a door is locked or in which direction a latch must be turned~\cite{wang2025adamanip}.
Opening a latched door, for instance, requires probing an unknown locked state or articulation direction through sequential attempts, accumulating evidence across interactions rather than acting on any single observation~\cite{li2025learn}.

Imitation learning has shown strong results in fully observable settings, where actions can be reliably mapped from recent observations~\cite{chi2025diffusion}.
A natural extension for partial observations is to condition the policy on the raw history of observations and actions~\cite{wang2025adamanip,zhang2025adaptive}.
However, history-conditioned policies face a fundamental difficulty in partially observable settings:
the same partial observation can be consistent with multiple hidden states that can only be differentiated based on interaction history. Different demonstrations may prescribe different actions for identical observation inputs.
This inconsistency degrades imitation learning performance in ways that additional training data alone cannot  resolve~\cite{li2025learn,setlur2025scaling,gangwani2020learning}.
In practice, such policies struggle to aggregate partial information across long episodes and lose track of earlier interactions as the horizon grows~\cite{luo2024multistage,zhou2025mtil,hong2024offline}.

We address this issue at the representation level by estimating the latent state history using Bayesian belief, similar to~\cite{kelestemur2022tactile}. Rather than providing raw history as policy input, we maintain a posterior belief $b_t = P(\mathbf{s} \mid z_{1:t})$ over the hidden state via sequential Bayesian filtering, and condition the policy directly on a compressed belief summary. This belief summary exposes two quantities the policy needs: the current best estimate of the hidden state, and the remaining uncertainty around that estimate.
This structured representation replaces the full observation history with a fixed-dimensional input that captures what the robot has inferred so far and how confident that inference is.
The uncertainty encoded in the belief guides the balance between exploratory, contact-seeking behavior and direct, goal-directed motion. This adaptive behavior requires no explicit mode switching, reward shaping, or auxiliary information-gain objectives. 

We evaluate BEACON in simulation across two domains: gripper alignment for sensor insertion and latched door opening. We compare against baselines that use only raw observations and privileged ground-truth state access.
In both domains, the belief-conditioned policy substantially outperforms the observation-only baseline and approaches privileged-state performance, with ablations confirming that the same trained policy adapts its level of probing to the initial belief uncertainty at inference time.

The contributions of this work are: (1) A belief-conditioned imitation learning framework for robot manipulation under partial observability, decoupling state estimation from policy learning; (2) Demonstration that belief conditioning enables a single trained policy to adapt between probing and committing based on the uncertainty in the current belief, without explicit mode switching or reward shaping; (3) Validation across two domains with qualitatively different belief representations, showing the approach is not tied to a specific form.

\section{Problem Formulation}~\label{sec:problem}
\vspace{-10pt}

 We formulate this as a POMDP defined by $(\mathcal{S}, \mathcal{A}, \mathcal{O}, T, Z)$, where $\mathcal{S}$ is the hidden state space, $\mathcal{A}$ the action space, $\mathcal{O}$ the observation space, $T$ the transition model (the identity here, as the hidden state is static and non-changing), and $Z$ the observation model. At each timestep the robot receives a partial, noisy observation $z_t \in \mathcal{O}$ that indirectly constrains $\mathbf{s}$. Because a single observation is generally insufficient to resolve $\mathbf{s}$, the robot must take a sequence of information-gathering actions to reduce uncertainty before committing to task completion. We maintain a belief $b_t = P(\mathbf{s} \mid z_{1:t})$ via Bayesian filtering, compressing the full history into a fixed-dimensional representation for decision-making. 
A learned policy $\pi(\mathbf{a}_t \mid \mathbf{o}_t, b_t)$ is conditioned on both the current observation and the updated belief, where $b_t$ exposes both the current state estimate and the remaining uncertainty. As the belief sharpens over an episode, the policy should transition from exploration (probing actions that elicit informative observations) to exploitation (committing to task completion based on a confident estimate). 
\section{Gripper Alignment via Belief Conditioning}~\label{sec:method}
\vspace{-20pt}

\begin{figure}[b]
    \centering
    \includegraphics[width=1\linewidth]{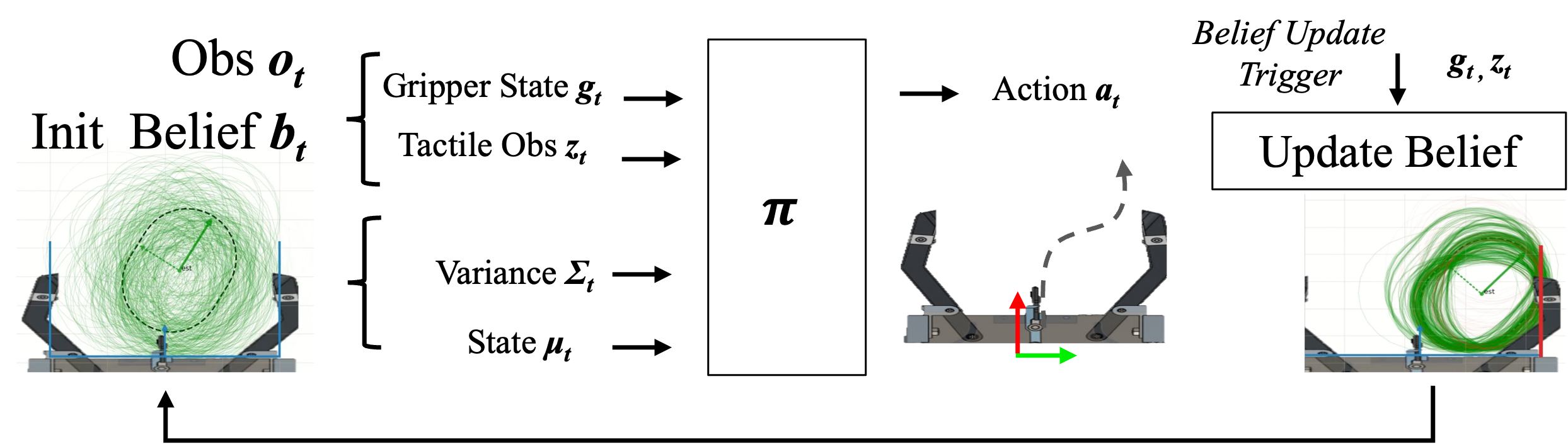}
    \caption{System diagram of the belief-conditioned policy. Varying belief guides the low-level policy.}
    \label{fig:method_overview}
    \vspace{-20pt}
\end{figure}

\subsection{Task Overview}
%describe task and environment.
This work is motivated by prior work on automating precise sensor insertion into plant stalks in field environments~\cite{lee2024towards}. Because visual estimation of stalk pose is often occluded by dense foliage, we improve pose estimation through tactile sensing. The robot makes a series of contacts, then commits to inserting the sensor at the center of the stalk along the surface normal.

%describe symbols
We simulate this task in 2D. The hidden stalk state is represented as a capsule-shaped object in 6D: $\mathbf{s} = (c_x,\; c_y,\; d,\; \cos 2\phi,\; \sin 2\phi,\; r)$, where $(c_x, c_y)$ is the capsule midpoint, $d$ is the separation between the two semicircle centers, $r$ is the radius, and $\phi$ is the capsule orientation. The double-angle encoding of $\phi$ resolves the $\pi$-periodic wraparound discontinuity and ensures a unique representation. The gripper pose is $\mathbf{g} = (g_x,\; g_y,\; \theta)$, with three line segments indexed $k \in \{0, 1, 2\}$ corresponding to the gripper left finger, palm, and right finger.

%%%%%%%%%%%%%%%%%%%%%%%%%%%%%%%%%%%%%%%%%%%%%%%%%%%%%%%%%%%%%%%%%%%%%%%%%%%%%%%%%%%%%%%%%%%%%%%%%%%%%%%%%
\subsection{Belief Representation}
%representation via Bayes rule, particle filter with manifold constraint (algo).
The belief over the hidden stalk state is maintained as a weighted particle set $\{\mathbf{s}^{(i)},\; w^{(i)}\}_{i=1}^{N}$, where each particle is a state hypothesis and its weight reflects the current posterior probability. At each step the robot receives a tactile observation $z \in \{\texttt{none},\; 0,\; 1,\; 2\}$ indicating which gripper side is in contact.

\begin{figure}[t]
    \centering
    \includegraphics[width=1\linewidth]{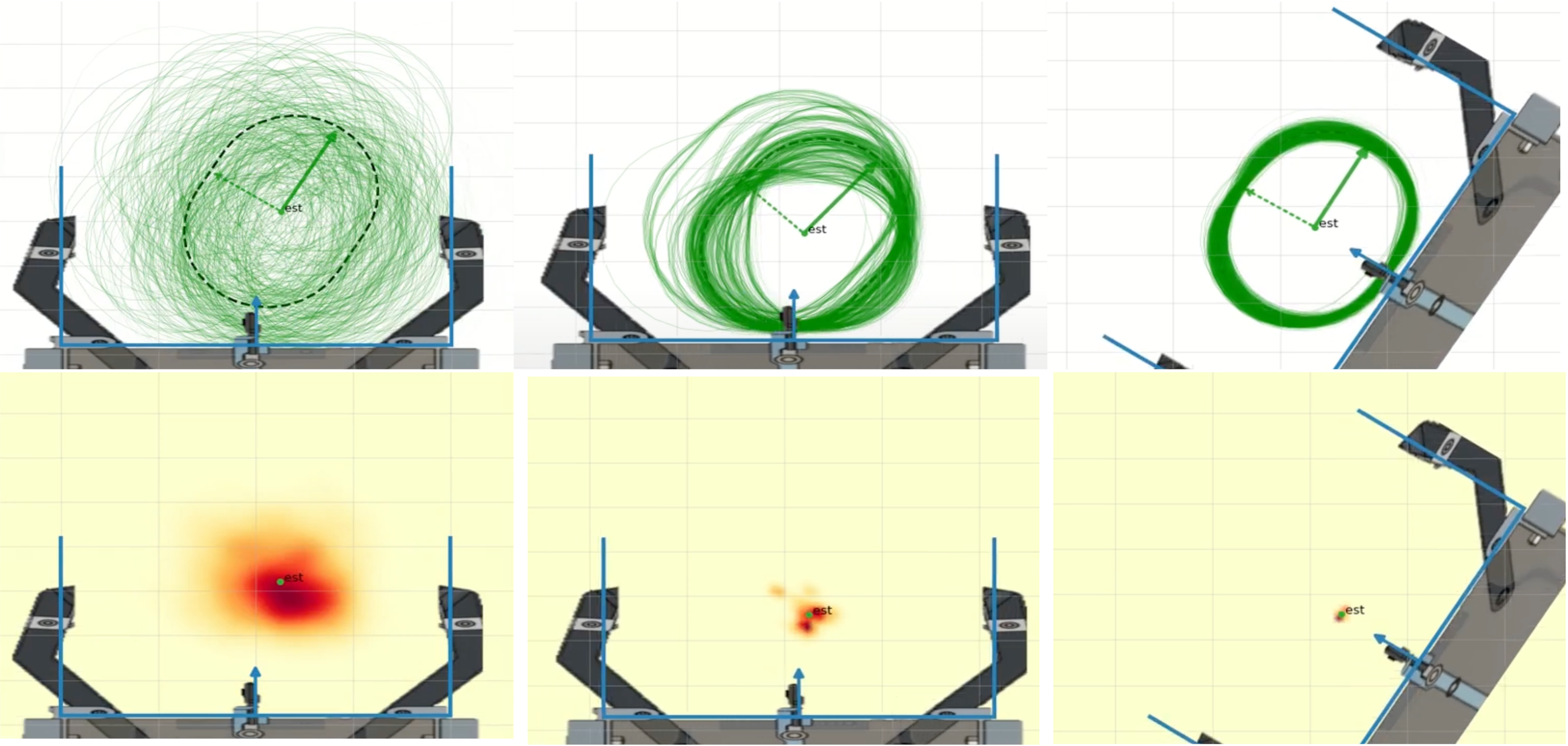}
    \caption{Sequential contact observations reducing uncertainty for gripper alignment.}
    \label{fig:uncertainty}
    \vspace{-15pt}
\end{figure}

By Bayes' rule, weights are updated multiplicatively and renormalized, $w_{\text{new}}^{(i)} \;\propto\; P(z \mid \mathbf{s}^{(i)}) \;\cdot\; w_{\text{old}}^{(i)}$. For each particle $i$, the minimum distance from the capsule to each gripper segment determines the nearest segment index $\hat{k}^{(i)}$ and corresponding distance $d_{\min}^{(i)}$. The soft contact probability is computed via a sigmoid over the signed penetration depth:

\begin{equation}
    \alpha^{(i)} = \sigma\!\left(\frac{r^{(i)} - d_{\min}^{(i)}}{\sigma_c}\right),
    \label{eq:alpha}
\end{equation}
where $\sigma_c$ controls the sharpness of the transition. This expression maps a probability that varies from $\alpha^{(i)} \approx 0$ (no contact) through $\alpha^{(i)} \approx 0.5$ (near contact) to $\alpha^{(i)} \approx 1$ (penetrated). The per-particle likelihood is then:
\begin{equation}
    P(z \mid \mathbf{s}^{(i)}) =
    \begin{cases}
        1 - \alpha^{(i)} & \text{if } z = \texttt{none}, \\[4pt]
        \alpha^{(i)} & \text{if } z = k \text{ and } \hat{k}^{(i)} = k, \\[4pt]
        \alpha^{(i)} \cdot \epsilon & \text{if } z = k \text{ and } \hat{k}^{(i)} \neq k,
    \end{cases}
    \label{eq:likelihood}
\end{equation}
where $\epsilon$ is a mismatch penalty applied when the particle predicts contact on the wrong segment.
Since standard particle filters are susceptible to particle starvation under sparse, discrete observations such as tactile contact, we adopt the Manifold Particle Filter (MPF)~\cite{koval2013manifold}. On contact, MPF supplements the conventional particle set by drawing additional samples directly on the contact manifold, improving coverage of high-likelihood regions.

%%%%%%%%%%%%%%%%%%%%%%%%%%%%%%%%%%%%%%%%%%%%%%%%%%%%%%%%%%%%%%%%%%%%%%%%%%%%%%%%%%%%%%%%%%%%%%%%%%%%%%%%%
\subsection{Dataset Collection}
%teleoperation
%input,output to train DP

Demonstrations are collected via human teleoperation in the 2D simulation environment for 50 demonstrations.
An interactive GUI renders the gripper, the particle-filter belief of the capsules, and the belief heatmap. The operator controls the gripper with keyboard commands mapping to $(\pm\Delta p_x, \pm\Delta p_y, \pm\Delta\theta)$.
An episode ends when the operator judges the gripper to be aligned and presses a key to deploy the sensor.

At each timestep $t$, the system records the tuple
$(\mathbf{o}_t,\; \mathbf{a}_t,\; \mathbf{s}_{\text{gt}})$.
The observation $\mathbf{o}_t$ consists of the gripper state
$\mathbf{g}_t = (g_x,\; g_y,\; \cos\theta,\; \sin\theta) \in \mathbb{R}^{4}$,
a discrete contact label $z_t \in \{-1, 0, 1, 2\}$ indicating which segment of gripper, and the full particle-filter belief
$\{\mathbf{s}^{(i)}_t, w^{(i)}_t\}_{i=1}^{N}$ with $N{=}1000$ particles in the 6D state space.
The action $\mathbf{a}_t = (\Delta p_x,\; \Delta p_y,\; \Delta\theta,\; e_t) \in \mathbb{R}^{4}$
contains the local-frame displacement command and a binary end-of-episode flag $e_t$.
The full $N$-particle cloud and weight vector are recorded at every step so that different
belief-compression strategies can be explored offline.
The concatenated observation vector
$\mathbf{o}_t = [\mathbf{g}_t;\; z_t;\; \mathbf{b}_t] \in \mathbb{R}^{32}$
and the action $\mathbf{a}_t \in \mathbb{R}^{4}$ are each min-max normalized to $[-1, 1]$.
Each training sample maps an observation
window $\mathbf{o}_{t:t+T_o}$ to a future action chunk $\mathbf{a}_{t:t+T_p}$, giving
$\mathcal{D} = \bigl\{
    (\mathbf{o}_{t:t+T_o},\; \mathbf{a}_{t:t+T_p})
    \;\big|\; t \in \mathcal{T}_{\text{valid}}
\bigr\}$, where $T_o=2$ and $T_p, T_a=4$.

%%%%%%%%%%%%%%%%%%%%%%%%%%%%%%%%%%%%%%%%%%%%%%%%%%%%%%%%%%%%%%%%%%%%%%%%%%%%%%%%%%%%%%%%%%%%%%%%%%%%%%%%%
\subsection{Belief-Conditioned Policy}

In a POMDP, a single observation can be consistent with many hidden states, making it difficult for end-to-end policies to disambiguate from raw observation history alone~\cite{li2025learn}. Our key insight is to leverage a Bayesian state estimator to maintain a structured belief that is refined with each contact, and to condition the policy directly on this belief. This allows the policy to reason about \emph{where uncertainty remains} and plan actions that both reduce it and progress toward gripper alignment. Because the task can be multimodal, where the gripper may need to approach from different directions depending on the current belief, we adopt diffusion policy~\cite{chi2025diffusion} to learn from human demonstrations this interactive task. Each observation $\mathbf{o}_t \in \mathbb{R}^{32}$ concatenates the gripper state, a contact label, and a compressed belief summary; each action $\mathbf{a}_t = (\Delta x_{\mathrm{loc}},\, \Delta y_{\mathrm{loc}},\, \Delta\theta,\, e) \in \mathbb{R}^{4}$ encodes a gripper motion and a binary end-of-episode flag.

\paragraph{Belief representation.}
Rather than passing the full $N$-particle cloud to the policy, we compress the belief into a Gaussian summary $\mathbf{b}_t \in \mathbb{R}^{27}$, consisting of the weighted mean $\boldsymbol{\mu}_t \in \mathbb{R}^{6}$ and the upper-triangular entries of the weighted covariance $\boldsymbol{\Sigma}_t \in \mathbb{R}^{21}$, computed as
$\boldsymbol{\mu}_t = \sum_i w^{(i)}_t \, \mathbf{s}^{(i)}_t$ and
$\boldsymbol{\Sigma}_t = \sum_i w^{(i)}_t (\mathbf{s}^{(i)}_t - \boldsymbol{\mu}_t)(\mathbf{s}^{(i)}_t - \boldsymbol{\mu}_t)^\top$. This compression is lossy but retains the first two moments of the posterior, which capture both the best current estimate and the remaining uncertainty across all six state dimensions. Both observations and actions are independently min-max normalized to $[-1, 1]$ on the training set. 

However, the Gaussian summary is inherently unimodal and cannot represent beliefs with multiple distinct hypotheses (e.g., two symmetric orientations equally plausible based on initial contact). To address this, we ablate an alternative
belief representation using a lightweight transformer encoder that maps the raw weighted particle cloud to a learned latent vector, trained end-to-end with the policy. Since the particles form an unordered weighted set, we use cross-attention conditioned on the current observation, allowing the encoder to produce a belief summary that adapts to the robot's current state rather than compressing the distribution in a fixed, context-independent manner.

\paragraph{Model architecture and training.}
The model starts from random noise and iteratively denoises to  action trajectory using a ConditionalUnet1D~\cite{chi2025diffusion} noise-prediction backbone. The $T_o$ observation is flattened into a conditioning vector, allowing the network to attend to belief changes across the observation window. Training follows the standard DDPM objective, minimizing the noise-prediction loss
$\mathcal{L} = \mathrm{MSE}(\boldsymbol{\epsilon},\; \boldsymbol{\epsilon}_\theta(\mathbf{o}_t,\; \mathbf{a}^k_t,\; k))$, where $\boldsymbol{\epsilon} \sim \mathcal{N}(\mathbf{0}, \mathbf{I})$
is the noise added at a randomly sampled diffusion step $k$.
At inference, actions are denoised from pure Gaussian noise over the full diffusion schedule, conditioned on the current observation window. After executing $T_a$ actions, the particle filter runs a full update cycle and the belief summary is recomputed for the next policy query.

\section{Experimental Results}~\label{sec:experiments}
We evaluate BEACON's ability to enable more adaptive behavior in a POMDP setting, specifically whether the policy explores when uncertainty is high and commits to task completion once uncertainty is sufficiently reduced. We test this hypothesis in the gripper alignment domain, comparing BEACON against baselines that use only raw observations (lower bound) or privileged ground-truth object state (upper bound).

\subsection{Gripper Alignment Domain}

\subsubsection{Baselines and Belief Representations}
We evaluate six configurations that vary the observation provided during both training and inference (Table~\ref{tab:belief_representations}). For metrics we use: success rate (model triggered sensor deployment and error is below threshold); episode length (Steps); and distance error (Dist) and alignment error (Align) measure sensor alignment precision.

\begin{table}[t]
    \centering
    \caption{Varying latent representations; from raw observations, to belief, to oracle.}
    \label{tab:belief_representations}
    \renewcommand{\arraystretch}{1.3}
    \scriptsize
    \resizebox{\columnwidth}{!}{%
    \begin{tabular}{ll cc ccc}
        \toprule
        & & \multicolumn{5}{c}{\textbf{Eval Metrics}} \\
        \cmidrule(lr){3-7}
        \textbf{Train Belief} & \textbf{Infer Belief}
          & SR\,(\%)$\uparrow$ & Ins\,(\%)$\uparrow$ & Steps$\downarrow$ & Dist\,(cm)$\downarrow$ & Align\,($^\circ$)$\downarrow$ \\
        \midrule
        Init Est.              & Init Est.\,(6D)        & 20  & 50           & 278         & 4.3          & 31.3 \\
        Belief $k{=}1.0$      & Belief\,(27D)          & 50  & 70           & 231         & 2.0          & 18.0 \\
        Belief $k{\in}[0,1]$ (ours) & Belief\,(27D)    & 80  & \textbf{100} & 139         & 1.4          & 11.9 \\
        Belief (transformer)      & Belief\,(128D)          & 80  & 90           & 137         & 2.3          & 17.5 \\
        Privileged             & Privileged\,(6D)       & \textbf{100} & \textbf{100} & \textbf{73} & \textbf{1.2} & \textbf{4.1} \\
        Privileged             & Init Est.\,(6D)        & 30  & 80           & 146         & 2.9          & 26.7 \\
        \bottomrule
    \end{tabular}%
    }
\end{table}

\begin{figure}[t]
    \centering
    \vspace{-10pt}
    \includegraphics[width=1\linewidth]{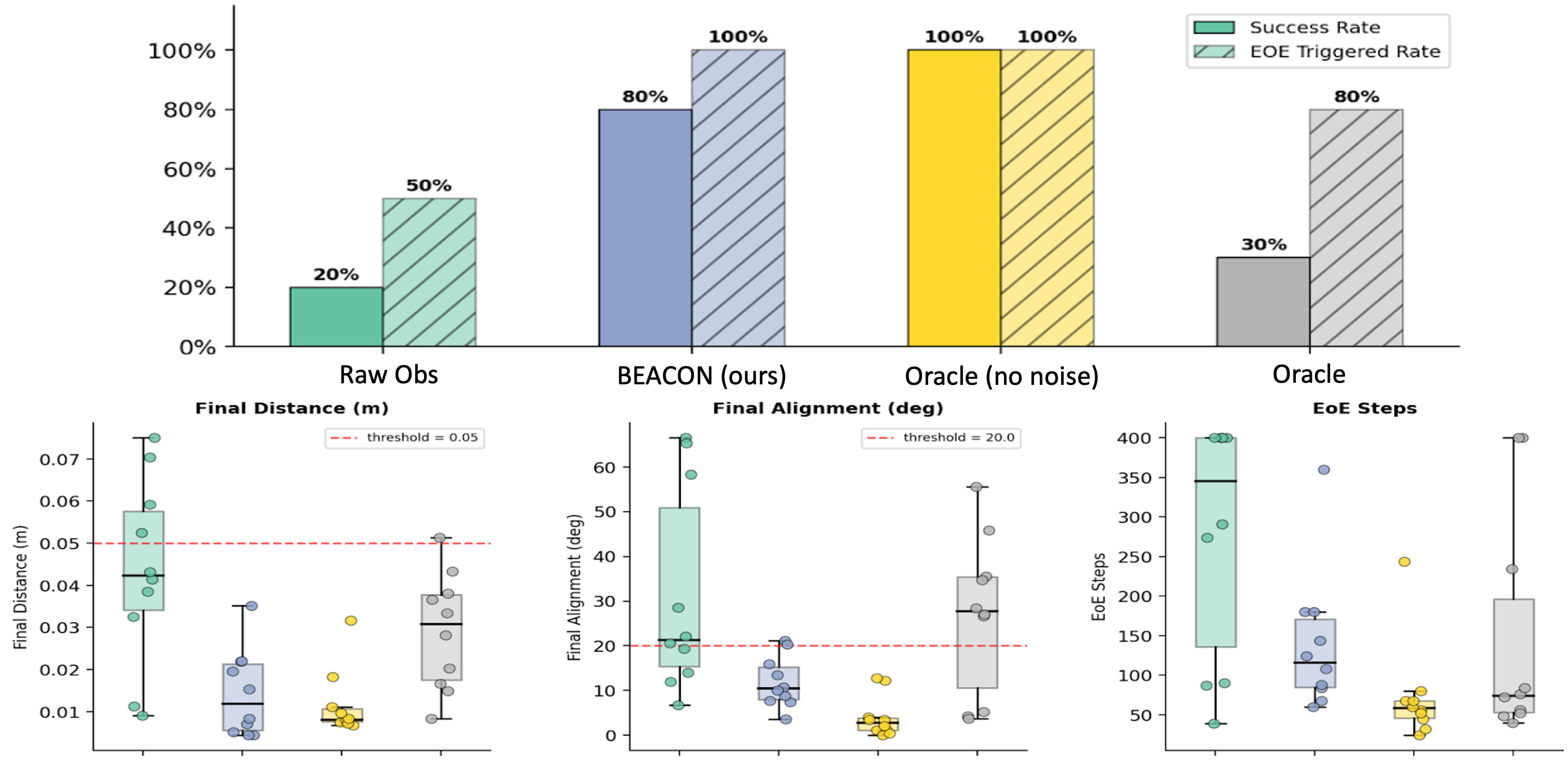}
    \caption{Performance comparison of varying policies. Raw observations (exp1) acts as lower bound, while oracle acts as upper bound (exp3). BEACON (exp2) reaches upper bound performance, while taking longer steps due to exploration.}
    \label{fig:performance}
    \vspace{-15pt}
\end{figure}

%baseline discussion
\textbf{No belief (lower bound)}. Config~1 provides only a fixed initial object estimate (e.g., a noisy, never-updated 6D pose) diffusion policy with observation history $T_o=2$. This mimics a setting where a visual prior is available but there is no state estimate. The core difficulty is that successful alignment requires a long sequence of spatially deliberate contacts. For this configuration, as latent state is not visually captured during demonstration, the state estimate is offloaded to the human operator to perform mentally by triangulating from previous observations. Without belief that aggregates past contacts into a refined estimate, the policy struggles on this long-horizon task: it takes the longest steps and has SR of only 20\%.

%changing initial estimate like visual servoing as the intial estimate changes. mean can quickly jump back-forth, which is why variance is needed (mean vs mean+variance)
\textbf{BEACON (proposed).} Config~2--4 replace the fixed estimate with a belief representation derived from the particle filter. Config~2 provides the full 27D Gaussian summary (mean and upper-triangular covariance) and achieves 50\% success in 231 steps. Config~3 additionally varies the initial noise level during training, exposing the policy to both high- and low-uncertainty demonstrations; this yields the best belief-conditioned result at 80\% success. The uncertainty terms are critical: without them, the policy cannot distinguish a confident estimate from an uncertain one, and tends to commit prematurely. 

\textbf{Privileged information (upper bound).} Config~5 conditions on the true hidden object state at both training and inference, achieving near-perfect performance. Demonstrations collected under this condition are entirely exploitative; the gripper aligns without exploration. Config~6 train with privileged information but evaluate using noisy estimates at inference, reveal the brittleness of this approach, achieveing only 30\% SR. Despite short episodes, the policy acts confidently on a noisy estimate and does not adapt to incoming tactile observations.

%%%%%%%%%%%%%%%%%%%%%%%%%%%%%%%%%%%%%%%%%%%%%%%%%%%%%%%%%%%%%%%%%%%%%%%%%%%%%%%%%%%%%%%%%%%%%%%%

\subsubsection{Ablation: Adaptive Exploration via Uncertainty Scaling}

A key question is whether the policy genuinely adapts its behavior to the current level of uncertainty. To test this, we manipulate the initial belief uncertainty at inference time and evaluate how the policy responds.

\textbf{Setup.} The 6D hidden object state is initialized with independent Gaussian noise. Because all noise dimensions are drawn independently, the prior covariance is diagonal with no cross-terms. We introduce a scaling factor  $k\in[0,1]$  applied to the standard deviations of this diagonal prior, which uniformly tightens or broadens the initial belief spread. At $k=1.0$ the belief is spread, reflecting high initial uncertainty; at $k=0.1$ the belief is tight, reflecting a confident initial estimate. We evaluate the same trained policy under $k \in \{0.1, 0.5, 1.0\}$ across 50 trials per condition, measuring episode length and tactile contact count as proxies for exploratory versus exploitative behavior.

\begin{figure}[t]
    \centering
    \includegraphics[width=1\linewidth]{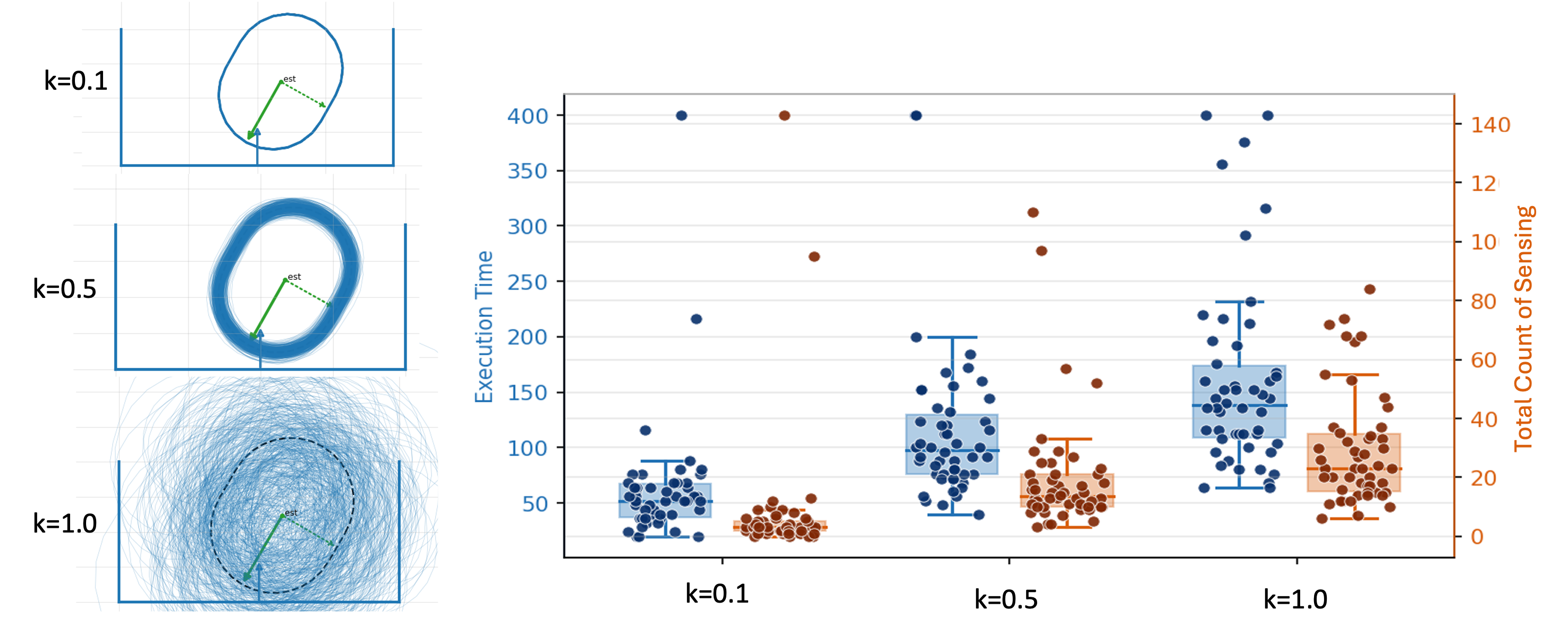}
    \caption{Ablation of exploration and exploitation based on uncertainty present.}
    \label{fig:ablation_noise}
    \vspace{-20pt}
\end{figure}

\begin{figure}[b]
    \centering
    \includegraphics[width=1\linewidth]{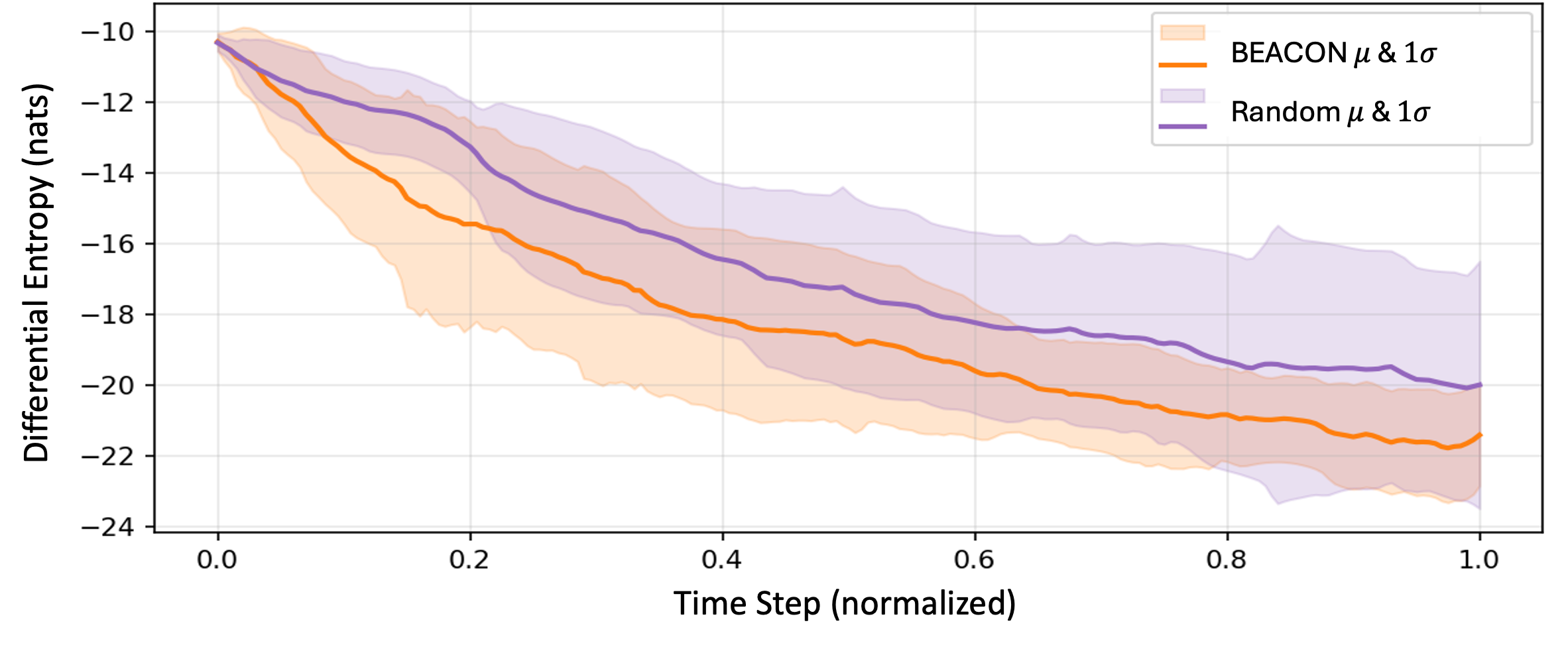}
    \caption{Uncertainty reduced using BEACON vs random directional search.}
    \label{fig:entropy}
    \vspace{-20pt}
\end{figure}

\textbf{Results.}
%ablation1: vary the covariance matrix in belief
 Under $k=0.1$, the policy produces short episodes with few tactile contacts, behaving exploitatively by committing quickly to the estimated object pose. Under $k=1.0$, the policy executes substantially longer episodes with a higher contact count before triggering insertion. The $k=0.5$ condition falls between these two extremes. This adaptive behavior is from the same policy, where the balance between exploration and exploitation is guided by the uncertainty encoded in the initial belief covariance. This property has a practical implication: the same policy can serve as both an exploratory and an exploitative agent depending on the quality of the initial state estimate, and without any retraining or explicit mode switching.

%additional metric: reducing uncertainty
\textbf{Implicit uncertainty reduction.}
We further examine whether the policy actively reduces belief uncertainty over the episode, despite having no explicit objective to do so. For the $k=1.0$ condition, we track the differential entropy of the belief, $h = \frac{1}{2}\ln\!\bigl((2\pi e)^d \lvert\Sigma\rvert\bigr)$, as a scalar measure of total uncertainty over time, and compare it against a random motion baseline that follows a billiard-like directional search, reversing or deflecting its heading upon contact. As shown in Fig.~\ref{fig:entropy}, the policy begins with high $nats$ and drives it steadily downward through targeted contact interactions, reaching  lower values by episode end than the random baseline achieves, as well as with lower deviation.

%%%%%%%%%%%%%%%%%%%%%% ADDING 2nd DOMAIN %%%%%%%%%%%%%%%%%%%%%%%%%%%%%%%%%%%%%%%%%%%%
%%%%%%%%%%%%%%%%%%%%%%%%%%%%%%%%%%%%%%%%%%%%%%%%%%%%%%%%%%%%%%%%%%%%%%%%%%%%%%%%%%%%%%%%%%%%%%%%%%%
%============================================================================================
\subsection{Latched Door Opening Domain}

\begin{table}[t]
    \centering
    \caption{Door domain: ablation across input features. Performance metrics mirror Domain~1; the decision-making metric captures interaction efficiency.}
    \label{tab:results}
    \renewcommand{\arraystretch}{1.3}
    \scriptsize
    \begin{tabular*}{\textwidth}{l@{\extracolsep{\fill}}l cccc}
        \toprule
        & & \multicolumn{3}{c}{\textbf{Performance}} & \textbf{Decision Making} \\
        \cmidrule(lr){3-5} \cmidrule(lr){6-6}
        \textbf{Model} & \textbf{Features}
          & SR $\uparrow$ & $T_{\text{exec}}$ $\downarrow$ & $\Delta\theta_H / \Delta\theta_L$ $\downarrow$
          & $N_{\text{grasp}}$ $\downarrow$ \\
        \midrule
        DP (baseline)        & $\mathbf{P}$              & 0.30          & 341          & 0.1 / 0.9 & 7.3 \\
        + tactile, history   & $\mathbf{P,T,H}$           & 0.55          & 254          & 0.1 / 0.1 & 3.5 \\
        + belief (BEACON)    & $\mathbf{P,T,H,B}$       & \textbf{0.85} & 225          & 0.1 / 0.0 & 3.3 \\
        + oracle             & $\mathbf{P,T,H,B,O}$     & 0.80          & \textbf{191} & 0.0 / 0.2 & 2.3 \\
        $-$ belief (oracle)  & $\mathbf{P,T,H,O}$       & 0.80          & 222          & 0.0 / 0.1 & 2.9 \\
        \bottomrule
    \end{tabular*}
    \vspace{-10pt}
\end{table}

\begin{figure}[h]
    \centering
    \includegraphics[width=1\linewidth]{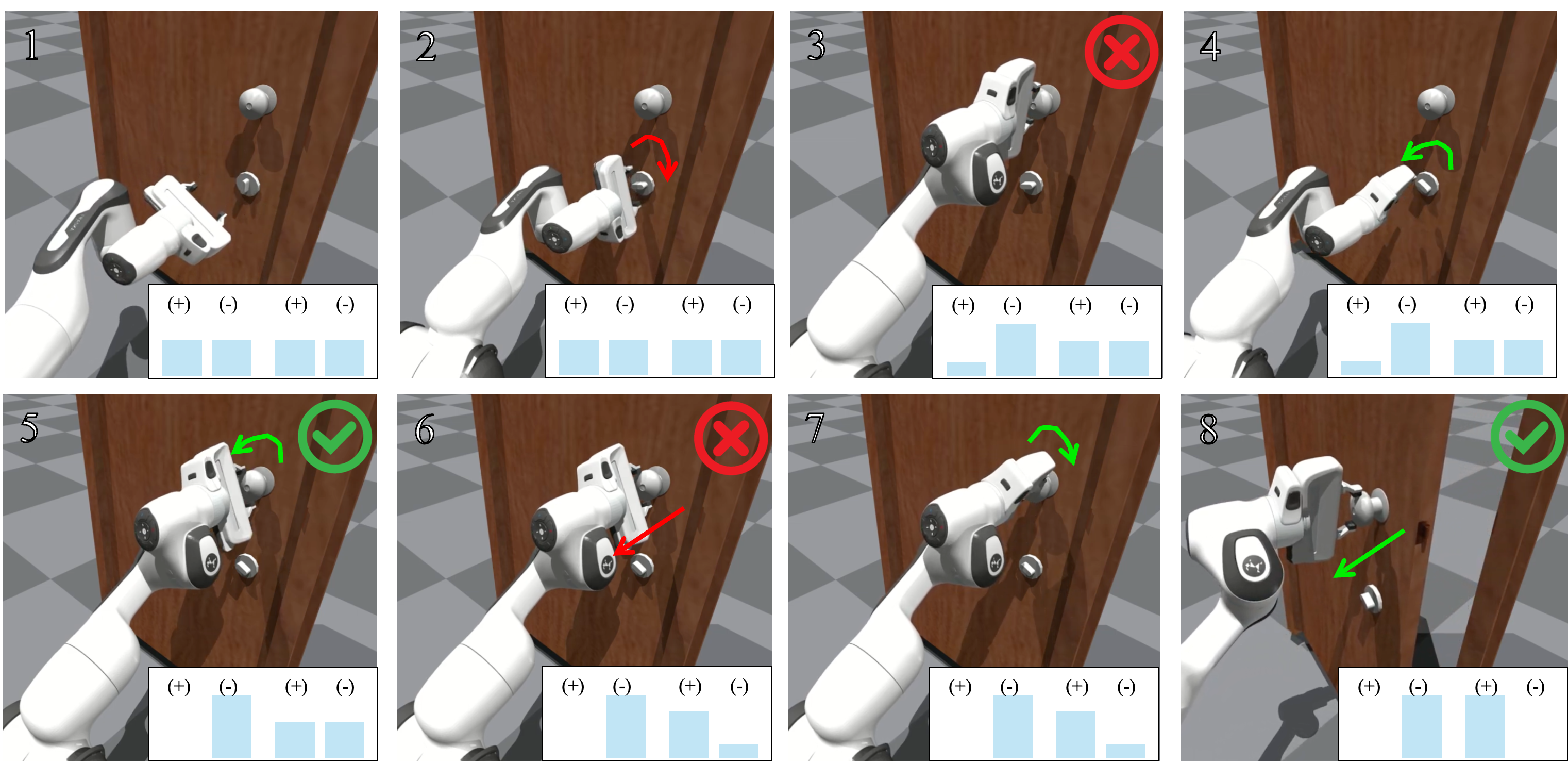}
    \caption{Belief over rotation direction for both latch and handle. Through observing the rotation of the locked handle and pulling on the door, the belief maintains estimate over true states to guide the policy.}
    \label{fig:door_belief}
    \vspace{-20pt}
\end{figure}

To test beyond the continuous-state setting, we apply BEACON to a different task needing a discrete distribution over categorical hypotheses.

\textbf{Task and hidden state.} A robot arm must open a latched door whose unlock mechanism is unknown: the latch must first be rotated past an unknown threshold $\bar{\theta}_L$, after which the handle must be rotated past a second unknown threshold $\bar{\theta}_H$, before the door can be pulled open. Each threshold is drawn from $[60^\circ, 90^\circ]$ at episode start with an unknown sign indicating rotation direction (CW or CCW). The robot observes only binary outcomes: whether a handle-rotation attempt succeeded ($o_H$) and whether a door-pull attempt succeeded ($o_D$). These outcomes reveal the latch and handle unlock states respectively, subject to  observation noise: false negatives occur with probability $\varepsilon$ (slippage causes a valid attempt to fail). Beyond the discrete hidden-state structure, the task adds long-horizon complexity through sequential interdependence (the handle cannot be probed until the latch is correctly set).

\textbf{Belief representation.} Latched door opening maintains a discrete categorical distribution over discretized hypothesis bins for each unlock threshold: $b_t(\hat{\theta}_L)$ and $b_t(\hat{\theta}_H)$. Beliefs are updated via Bayes' rule using the asymmetric observation likelihood
\begin{align}
    P(o_H = 1 \mid \theta_L,\, \hat{\theta}_L) &= U_L(\theta_L,\, \hat{\theta}_L) \cdot (1 - \varepsilon), \label{eq:obs_contact} \\
    P(o_H = 0 \mid \theta_L,\, \hat{\theta}_L) &= 1 - U_L(\theta_L,\, \hat{\theta}_L). \label{eq:obs_no_contact}
\end{align}
where $U_L \in \{0,1\}$ indicates whether the commanded latch angle exceeds the hypothesized threshold. The handle belief is updated analogously from $o_D$. Both discrete beliefs are concatenated with the gripper state and recent observation history to form the policy's conditioning input, mirroring Domain~1's framework.

\textbf{Baselines and configurations.} We evaluate five model configurations that progressively augment the diffusion policy's input features (Table~\ref{tab:results}): proprioception alone ($\mathbf{P}$), with tactile and history ($\mathbf{P,T,H}$), with the proposed belief ($\mathbf{P,T,H,B}$), with hidden oracle state ($\mathbf{P,T,H,B,O}$), and an ablation removing belief from the oracle setting ($\mathbf{P,T,H,O}$). Episodes are scored on four metrics. Performance metrics mirror Domain~1: success rate (SR, door opens past $\theta_D^{\text{open}}$), execution time ($T_{\text{exec}}$, capped at 500 steps), and unlock error ($\Delta\theta_H, \Delta\theta_L$, the residual angle to each threshold at termination). The decision-making metric captures interaction efficiency through the number of distinct grasp events ($N_{\text{grasp}}$), which count total number of grasp of door handle and latch across the episode. Higher count could indicate more exploratory behavior or mindless grasp attempts.

\textbf{No belief (lower bound).} The proprioception-only baseline (DP) achieves only 30\% success with the highest grasp count (7.3) and longest execution time (341 steps), reflecting near-arbitrary motion through the demonstration distribution without a coherent notion of mechanism state. Adding tactile and history ($\mathbf{P,T,H}$) lifts SR to 55\% and roughly halves the grasp count, suggesting the policy begins to track interaction outcomes implicitly; without a structured belief, it cannot reliably aggregate past observations into a stable estimate of the unlock thresholds.

\textbf{BEACON (proposed).} Conditioning on the discrete belief ($\mathbf{P,T,H,B}$) achieves 85\% success in 225 steps with low residual unlock error ($\Delta\theta_H = 0.1$, $\Delta\theta_L = 0.0$), matching the privileged oracle's performance within the variability of the trial budget. The policy reads the belief over $\hat{\theta}_L$ and $\hat{\theta}_H$ to determine when each unlock condition has been confidently established, then commits to the door-pull phase. As in Domain~1, this requires somewhat longer episodes than the oracle, since the policy must gather observations to refine the belief rather than acting on ground truth directly.

\textbf{Privileged information (upper bound).} Conditioning on oracle threshold values ($\mathbf{P,T,H,B,O}$) reaches 80\% SR in 191 steps, the fastest of all configurations. Removing the belief from this configuration ($\mathbf{P,T,H,O}$) drops SR by 5 points and raises $\Delta\theta_L$, indicating that even with privileged thresholds available, the structured belief contributes useful signal about action-outcome consistency.

\textbf{Analysis.} The progression from baseline to belief-conditioned policy parallels Domain~1's pattern: raw observations are insufficient, the belief representation closes most of the gap to the privileged upper bound, and final precision improves as the policy gains access to a structured estimate of the hidden state. The key takeaway is that this pattern holds across two qualitatively different belief representations: a Gaussian summary of a continuous particle cloud and a discrete categorical distribution over hypothesis bins. Belief conditioning is not specific to a particular density form; what matters is that the conditioning input expose both the current estimate and the uncertainty around it, in whatever representation the underlying state space requires.

\section{Conclusion}

We presented BEACON, a belief-conditioned imitation learning framework for robot manipulation under partial observability.
Rather than conditioning on raw observation history, we maintain a Bayesian belief over the hidden state and condition a diffusion policy on a compressed summary that exposes both the current estimate and remaining uncertainty.
This structured input enables a single trained policy to implicitly balance exploration and exploitation based on belief uncertainty, without explicit mode switching or reward shaping.
\textbf{Limitations and future work.}
The current framework has two primary limitations.
First, the observation model $P(z \mid \mathbf{s})$ is hand-crafted per domain, requiring manual engineering of sensor likelihoods and limiting applicability to tasks where contact geometry is well understood.
Second, the hidden state space must be pre-specified: its structure, dimensionality, and parametrization are defined in advance rather than discovered from data.
Both limitations require substantial domain knowledge before any learning can take place, and therefore limits the generalization of this method. 
Future work should address these by learning the observation model directly from interaction data, and by replacing the pre-specified state representation with a latent embedding discovered through self-supervised or contrastive objectives over contact sequences.

\bibliographystyle{splncs04}
\bibliography{mybib}

\end{document}